\documentclass{article}

\usepackage[utf8]{inputenc}
\usepackage[T1]{fontenc}
\usepackage{hyperref}
\usepackage{url}     
\usepackage{booktabs} 
\usepackage{nicefrac}       
\usepackage{microtype}
\usepackage{xcolor}         
\usepackage{cite}
\usepackage{amsmath,amssymb,amsfonts}
\usepackage{algorithmic}
\usepackage{graphicx}
\usepackage{textcomp}
\usepackage{makecell}
\usepackage{tikz}
\usepackage{tabularx}
\usepackage{float}
\newcolumntype{C}{>{\centering\arraybackslash}X}
\newcolumntype{R}{>{\raggedleft\arraybackslash}X}
\newcolumntype{L}{>{\raggedright\arraybackslash}X}
\usepackage{subcaption}
\usepackage[numbers]{natbib}

\usepackage[preprint]{neurips_2024}

\title{\emph{WarCov} -- Large multilabel and multimodal dataset from social platform}

\author{
  Weronika~Borek-Marciniec \\
  \texttt{weronika.borek@pwr.edu.pl} \\
  \And
  Paweł~Zyblewski \\
  \texttt{pawel.zyblewski@pwr.edu.pl} \\
  \And
  Jakub~Klikowski \\
  \texttt{jakub.klikowski@pwr.edu.pl} \\
  \And
  Paweł~Ksieniewicz \\
  \texttt{pawel.ksieniewicz@pwr.edu.pl} \\
  \AND\vspace{-1.4em}
  ~\\
  Wroclaw University of Science and Technology, Department of Systems and Computer Networks \\
  Wyb. Wyspiańskiego 27, 50-370 Wrocław, Poland \\
}

\begin{document}

\maketitle

\begin{abstract}
    In the classification tasks, from raw data acquisition to the curation of a dataset suitable for use in evaluating machine learning models, a series of steps -- often associated with high costs -- are necessary. In the case of \emph{Natural Language Processing}, initial cleaning and conversion can be performed automatically, but obtaining labels still requires the rationalized input of human experts. As a result, even though many articles often state that "\emph{the world is filled with data}," data scientists suffer from its shortage. It is crucial in the case of natural language applications, which is constantly evolving and must adapt to new concepts or events. For example, the topic of the COVID-19 pandemic and the vocabulary related to it would have been mostly unrecognizable before 2019. For this reason, creating new datasets, also in languages other than English, is still essential. This work presents a collection of 3~187~105 posts in Polish about the pandemic and the war in Ukraine published on popular social media platforms in 2022. The collection includes not only preprocessed texts but also images so it can be used also for multimodal recognition tasks. The labels define posts' topics and were created using hashtags accompanying the posts. The work presents the process of curating a dataset from acquisition to sample pattern recognition experiments.
\end{abstract}

\section{Introduction} 
\label{sec:intro}
It can be considered paradoxical that many machine learning articles begin with the statement, "\emph{The world is filled with data}.", while at the same time, researchers point out the shortage of real-world datasets on which to perform research~\cite{manco2022machine}. In many fields, it results in generators based on existing datasets~\cite{lu2023machine} but also brings risks, especially in \emph{Natural Language Processing}, where models learned on data artificially generated using other models (or even their previous versions) tend to degrade the acquired generalization ability~\cite{zhang2021commentary}.

There is no clear consensus about image data. On the one hand, methods are emerging to improve recognition quality using generated data~\cite{besnier2020dataset}. On the other hand, it is pointed out that multiple inbreeding looping of the learning process on the own output results in an increasing degradation in the generalization capabilities of the model~\cite{Hataya_2023_ICCV}.

Jamain et al.~\cite{jamain2009large} conducted a thorough analysis of this topic already in 2009, showing a shortage in real datasets that were both large and difficult. Unfortunately, a more up to date analysis has not been done since, but the literature points out that the data used to train models often stays unpublished, and thus, does not provide a way to replicate the research~\cite{kapoor2023leakage}. Carlini et al.~\cite{carlini2021extracting} have developed a method for training data extraction attacks, with which it is possible to recover some of the data that was used to build the model in an almost unchanged form. This phenomenon is called regurgitation by large model developers. This type of data may contain sensitive information, such as names or phone numbers, which is probably one of the reasons why the collections are not published. This also indicates a scale of carelessness in the processing and cleaning training data, as there are sets of directives specifying the data's anonymization~\cite{9298747}.

Using content published on the Internet to train machine learning models is being attempted to curb not only by companies but also by creators themselves, if only by processing their own graphics with dedicated programs designed to make training the model on them cause a decrease in recognition quality~\cite{shan2023glaze}. However, this is still an open debate, in which the comfort of the big players in extracting significant money from their solutions is most likely to prevail. Additional problems are indicated for linguistic data. Natural language is dynamically changing, which can cause quality deterioration over several years if the models are not updated~\cite{jaidka2018diachronic}. However, the biggest problem is more glaring -- as is known, there are many languages worldwide. Many multi-language models have been proposed, with good recognition quality achieved through transfer learning. Nevertheless, most are trained in at least part corpora from all the languages included in the model~\cite{kassner2021multilingual}.

Consequently, the fact that most available datasets are in English does not prevent the need to acquire and publish datasets in less popular languages. Therefore, this paper proposes a \emph{WarCov}: both described and openly implemented procedure of preprocessing and its realisation as a multimodal dataset consisting of preprocessed (a) posts published on a well-known social media platform in Polish and (b) photos and graphics attached to these posts.

Unlike the most of available data, presented dataset is dedicated to classification tasks, not text generation, for which at least the popular BERT language models were originally intended~\cite{devlin2018bert}. \emph{WarCov} has been prepared in such a way as to allow for the most reliable benchmarking of artificial intelligence methods in at least several specific recognition tasks. The leading one here is multilabel classification because each object in the set is assigned to at least one of fifty categories. The dataset also provides a proper basis for research in the task of clustering linguistic material, where the provided embeddings can serve as a representation in which we look for the optimal division, taking into account the possibility of belonging to more than one cluster. The aspect of multimodality, which, thanks to the available image data, supplements the linguistic set with several tens of thousands of matched text-image pairs, allows for research in the field of knowledge transfer between modalities. The sufficient size of the image pool itself also allows for research solely related to computer vision.

In brief, the main contributions of this article are as follows:

\begin{itemize}
    \item Providing access -- in the form of standardized embeddings -- for the machine learning community to the \emph{multilabel} and \emph{multimodal} dataset \emph{WarCov} under CC BY-NC-SA 4.0 license~\footnote{\url{https://creativecommons.org/licenses/by-nc-sa/4.0/}}~\footnote{Upon request on the project's GitHub page, we declare that we will extract features from the original dataset using any indicated method if possible to run with the computational resources of our University. The obtained embeddings will be added to a publicly available data repository in such a case. We also encourage researchers to fork the solution for their projects.}, where:
    \begin{itemize}
        \item Textual data are provided as word embeddings obtained using a language mode \emph{clips/mfaq}~\cite{debruyn2021mfaq} -- based on the XLM-RoBERTa~\cite{conneau-etal-2020-unsupervised} and additionally trained on multilingual short texts. 
        \item The images are provided as (i) embeddings extracted from the raw dataset using the ResNet-18 architecture pre-trained on the ImageNet dataset and (ii) embeddings extracted for 80\% of the set using the ResNet-18 architecture pre-trained on the ImageNet dataset and finetuned on the remaining 20 \% of images.
        \item At the end texts and images preprocessed with \textsc{pca} to keep the same number of components. 
    \end{itemize}
    \item A precise description of the procedure of acquisition and standardized preprocessing of the set -- taking into account the adopted criterion for obtaining labels -- in the form of an open-sourced \verb|@w4k2/warcow| project published on the GitHub platform\footnote{\url{https://github.com/w4k2/warcow}} (Section~\ref{sec:dataset}).
    \item Experimental evaluation validating the WarCov dataset's complexity as a benchmark in the multilabel classification task (Section~\ref{sec:exps}).
\end{itemize}

\section{Related works} 

Real-world datasets are vital when designing machine learning solutions with practical applications in mind~\cite{sarker2021machine}. One of the most popular sources of a real content is social media, especially microblogs, which provides compact, but rich and wide in sampling of users opinion on a given topic, allowing for the preparation of reliable datasets~\cite{10.1145/1963405.1963500}. Such medium do provide a wide range of possibilities for developing datasets with statements describing various phenomena, such as presidential elections~\cite{chen2022election2020}, the COVID-19 epidemic~\cite{lamsal2021design}, climate change~\cite{EFFROSYNIDIS2022117541}, cyberbullying~\cite{van2020multi}, or disinformation~\cite{HAYAWI202223}. The shared content is often conversational, allowing more specialized datasets for discovering various sentiments to be designed~\cite{xiong-etal-2019-tweetqa}.

Recent research shows that using multimodal approaches is more beneficial than focusing processing on a single modality~\cite{10041115}. Due to the above, the need for access to real multimodal sets is also increasing. In the literature, a frequently discussed problem of multimodal classification of social media data is fake news detection~\cite{QU2024102172}. One well-known set is Fakeddit~\cite{nakamura-etal-2020-fakeddit}, which covers this topic based on data from the Reddit platform. An interesting example is also the \textsc{alone} dataset~\cite{wijesiriwardene2020alone}, in which the authors provide multimodal data for the problem of recognizing toxic behavior on Twitter. Equally important among the real datasets is the availability of multiple labels. For example, the \emph{NELA-GT-2020}~\cite{gruppi2021nelagt2020} set, and its predecessors~\cite{Norregaard_Horne_Adali_2019}, is a collection of newspaper articles prepared for a misinformation detection problem.

Many datasets that appear in the literature are limited to English. Current trends in Natural Language Processing strongly underline the multilingual aspect of the proposed methods~\cite{nguyen-etal-2021-trankit}. It creates an additional need to acquire real data in a broader linguistic spectrum. Factify~\cite{suryavardan2023factify} is a multimodal dataset of 100,000 English and Hindi articles marked for fact-checking and satire. MMChat~\cite{zheng-etal-2022-mmchat} includes a dataset of multimodal conversations from Chinese social media. Bondieli et al.~\cite{BONDIELLI2024110440} propose a set of fake news detections in Italian. LUMINA~\cite{SETYANINGSIH2024110279} is an Indonesian multimodal set containing audio and video to support research in speech perception.

A specific part of the multilingual datasets are works based on texts written in Polish. It is possible to find unimodal datasets, a benchmark of \textsc{nlp} tasks~\cite{rybak-etal-2020-klej}. This collection of sets includes the CBD set~\cite{ptaszynski2019results} for automatic cyberbullying detection on Polish social media. Similarly, there are separate sets for the content summarization task~\cite{ogrodniczuk-kopec-2014-polish} and the Question Answering task~\cite{marcinczuk2013open}, both in Polish. There are very few multimodal sets with Polish content. One example is the POLEMAD set~\cite{WIELGAT202129}, which contains multimodal speech data for the Polish language obtained from an electromagnetic articulograph. To our knowledge, there is currently no multimodal and multilabeled dataset for Polish based on social media data.

\section{Dataset}
\label{sec:dataset}

The dataset consists of microblog posts published by users of the popular social media platform in 2022. Raw data was acquired in January and February 2023 using the approved, official \textsc{api} with academic access. It should be taken into account that some of the posts may no longer be available due to deletion by the author or the platform itself. The data structure is described in Subsection~\ref{subsec:data_structure}.

Posts were searched using a language filter (Polish) and four keywords: \verb|covid|, \verb|szczepionki|, \verb|szczepienia| (eng. \emph{vaccinations}) and \verb|Ukraina| (eng. \emph{Ukraine}). Predicate was in constructed in order to collect posts regarding two internationally essential, but country specific events -- the global \textsc{covid}-19 pandemic and Russia's aggression against Ukraine.

\subsection{Data structure} 
\label{subsec:data_structure}

\begin{figure}[!htb]
    \centering
    \includegraphics[width=.99\textwidth]{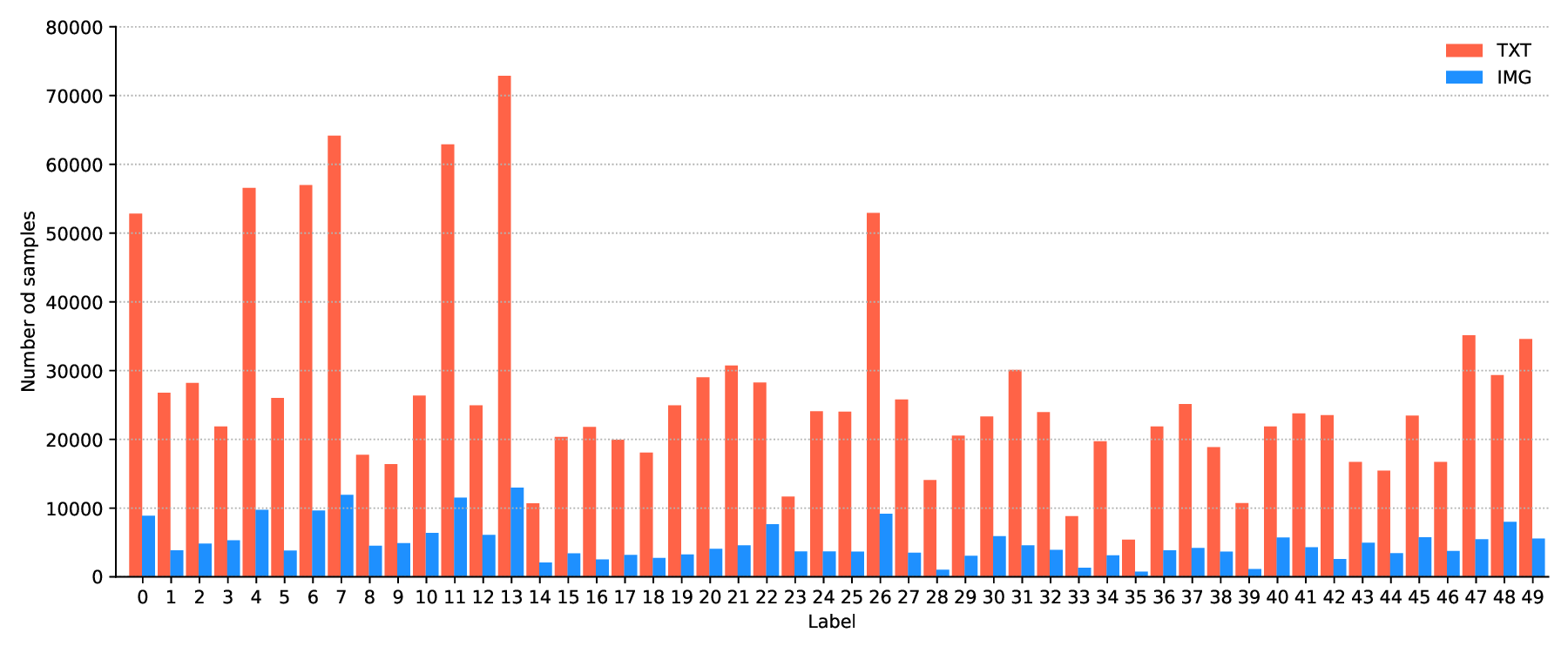}
    \caption{Label distribution for text and image modalities}
    \label{fig:img}
\end{figure}

The dataset consists of 3~187~105~microblog post objects. The objects are represented as (a) post text, (b) publication date, (c) image, if post contained it and (d) a collection of 50 binary labels. Additionally, 588~046 of them ($\sim18.5\%$) were also hashtagged, and 87~816 ($\sim3\%$) contained images (Figure~\ref{fig:img}). 

\subsection{Labels preparation} 
\label{subsec:labels_prep}

The dataset is intended for a classification task, where, based on the post content and/or image, patterns are assigned a label reflecting the topic the object concerns. Labels for the classification task were prepared using hashtags attached to posts. Their deep representation was produced using the model \emph{clips/mfaq}~\cite{debruyn2021mfaq} with \emph{Sentence Transformer}~\cite{reimers-2019-sentence-bert}.

Models used to build embeddings return objects represented by hundreds of features, significantly increasing the representation size. This results in increased learning time for the model and the need to provide more storage space. For this reason, to make the data highly accessible, it was decided to analyze the explained variance of deep representation and compress it using \emph{Principle Component Analysis} (\textsc{pca}), which made it possible to determine that 80\% of the described variance is concentrated in primary 39 components out of the existing 768. Representation of the hashtags was supplemented by date of publication (day, month, hour, minutes) to consider take into account potential \emph{concept drift}~\cite{bechini2021addressing}.  

The number of unique hashtags was too high to use them as a proper source of informative bias when unprocessed, so an additional strategy for obtaining cluster-based labels was developed. The biggest challenge of this stage was selecting the target number of possibly homogenous clusters to describe the problem. A human expert's analysis of the clustering runs for different counts was made to select the threshold of 50. Such a value, on the one hand, provided a fair variety of topic groups that would allow sufficiently distinctive hashtag assignment to a group and, on the other -- to create a problem in which categories are represented by enough count to make them possible to distinguish over a level of a random classifier.

Clustering was performed using the \emph{KMeans} method, combining processing speed with experimentally proven quality. It is also suitable for text embeddings because it belongs to the group of minimal-distance methods~\cite{shashavali2019sentence}. Self-learning paradigm was used to propagate labels to objects that initially did not contain hashtags. To do it, an inductive clustering model was used to make a prediction for unhashtagged samples. Due to automatic labeling, it was decided to use only objects originally containing hashtags in the experiments described in Section~\ref{sec:exps}.

\begin{figure}[!htb]
    \centering
    \includegraphics[width=.5\textwidth]{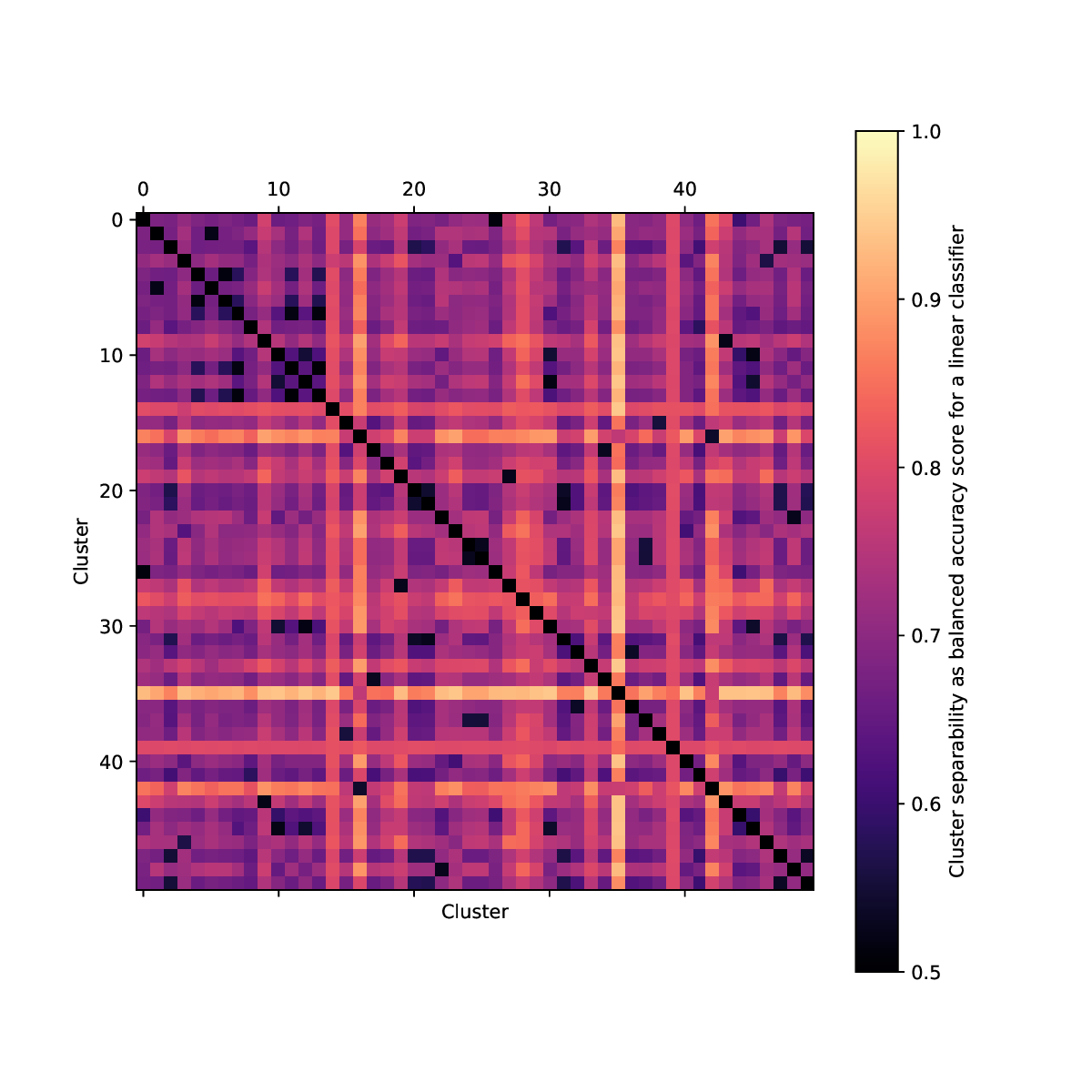}%
    \includegraphics[width=.5\textwidth]{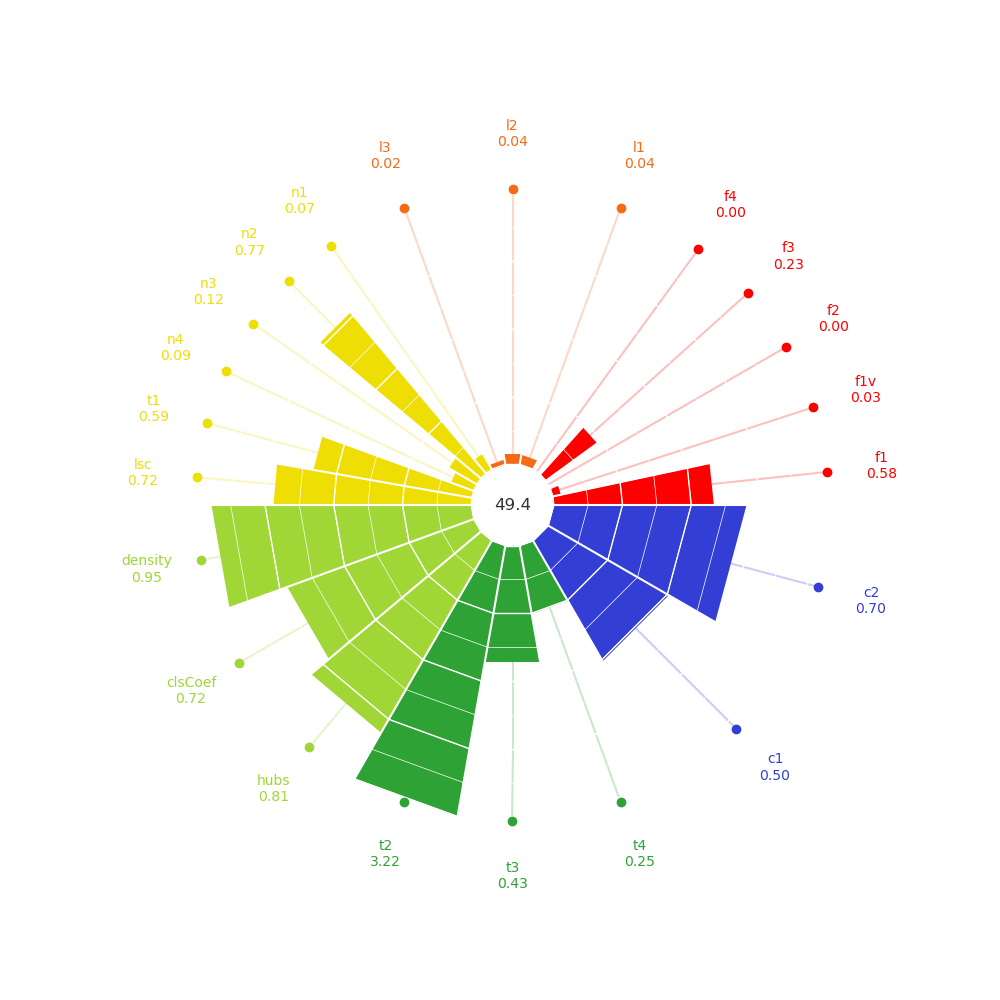}
    \caption{Separability of the obtained categories (left) and problem complexity measures (right)}
    \label{fig:diff-array}
\end{figure}

Separability of the obtained categories may be verified with simple experiment visualized in the left side of Figure~\ref{fig:diff-array}. A simple linear model was build between each pair of classes achieved due to the described procedure and validated using \emph{balanced accuracy score}. Achieved quality shows that separability varies from classes easy to distinguish, to the hard cases. The right side of Figure~\ref{fig:diff-array} presents the overall problem complexity of a dataset, as calculated by \emph{problexity} library~\cite{komorniczak2023problexity, lorena2018complex}.

\subsection{Preprocessing}
\label{subsec:prepro}

The first step in preprocessing was the preparation of text data. To properly process the texts with the classification model, it is necessary to perform text vectorization. The \emph{clips/mfaq}~\cite{debruyn2021mfaq} model was used to obtain a suitable representation based on XLM-RoBERTa~\cite{conneau-etal-2020-unsupervised} pre-trained with a multilingual set of FAQs. The ability to process the Polish language and the adaptation to process short natural language content makes this model perfectly suited for the microblog data.

Although images are paired with only 3\% of the collected texts, due to the overall large size of the dataset (3~187~105 texts), it is possible to extract a potentially valuable subset of multimodal data containing 87~816 samples represented by both text and image. Due to the need to anonymize the collected data, images can only be published as embeddings extracted using a convolutional deep neural network. For this purpose, the ResNet-18~\cite{he2016deep} architecture pre-trained on the ImageNet~\cite{deng2009imagenet} dataset was used as a base. The process of preparing available image data for processing consisted of two standard steps:

\begin{enumerate}
    \item \textbf{Analysis of the available color channels:} ResNet-18 accepts digital images containing only three color channels. Therefore, in RGBA images, we leave only RGB channels, and in greyscale images, we duplicate the only channel.
    \item \textbf{Image preprocessing using modified inference transforms:} Using ResNet-18 pre-trained on ImageNet, images should be resized to $256\times 256$ px using bilinear interpolation and then centrally cropped to $224\times 224$ px, followed by rescaling the pixel values to $[0.0, 1.0]$ and then normalizing them using $mean=[0.485, 0.456, 0.406]$ and $std=[0.229, 0.224, 0.225]$. In our case, due to the desire to keep full images for fear of losing valuable information that may be found, for example, on banners of news programs, the images were initially resized to $224\times 224$ instead of resizing them to $256\times 256$ px and then cropping.
\end{enumerate}

In all the experiments presented in this article and the feature extraction process, ResNet-18, pre-trained on ImageNet data, was used~\cite{paszke2017automatic}. In cases of fine-tuning, \emph{Stochastic Gradient Descent} with learning rate $0.001$ and momentum $0.9$ was used as the optimizer, the loss function was defined as \emph{multiLabel Soft Margin Loss}, and the batch size for each epoch was set as 8.

\section{Preliminary experiments}
\label{sec:exps}

The presented WarCov dataset can be used as a benchmark for the multilabel classification of text, images, and multimodal data. In order to illustrate the usefulness of \emph{WarCov}, sample preliminary experiments were conducted, offering insight into the generalization ability achieved by sample pattern recognition methods for each of the mentioned tasks. The classification methods were implemented with \emph{scikit-learn} library~\cite{scikit-learn}. All results were obtained using the 5 times repeated stratified 2-fold cross-validation ($5 \times 2$ \textsc{cv})~\cite{stapor2021design} protocol and can be replicated using the code available in the \emph{GitHub} repository of the project~\footnote{\url{https://github.com/w4k2/warcow}}. The experiments were conducted using Mac Studio with Apple M1 Ultra with 20-core CPU, 64-core GPU, 32-core Neural Engine system, and 128 GB RAM.

\subsection{Mutli-label text classification}

The experiments for the text modality were conducted using all instances that originally contained hashtags, so their labels are the result of direct propagation and not generation by the model. Due to the already-mentioned tendency of language models to create high-dimensional embeddings, the \textsc{pca} method was also used to limit the number of features. It was decided to indicate a threshold of 95\% of the variance -- more than in case of only hashtags, as there is a significant difference between length of the single word or phrase and the whole post. It was more crucial to keep as many information as possible. 

\begin{figure}[!htb]
     \centering
     \begin{subfigure}[b]{0.49\textwidth}
         \centering
         \includegraphics[width=\textwidth]{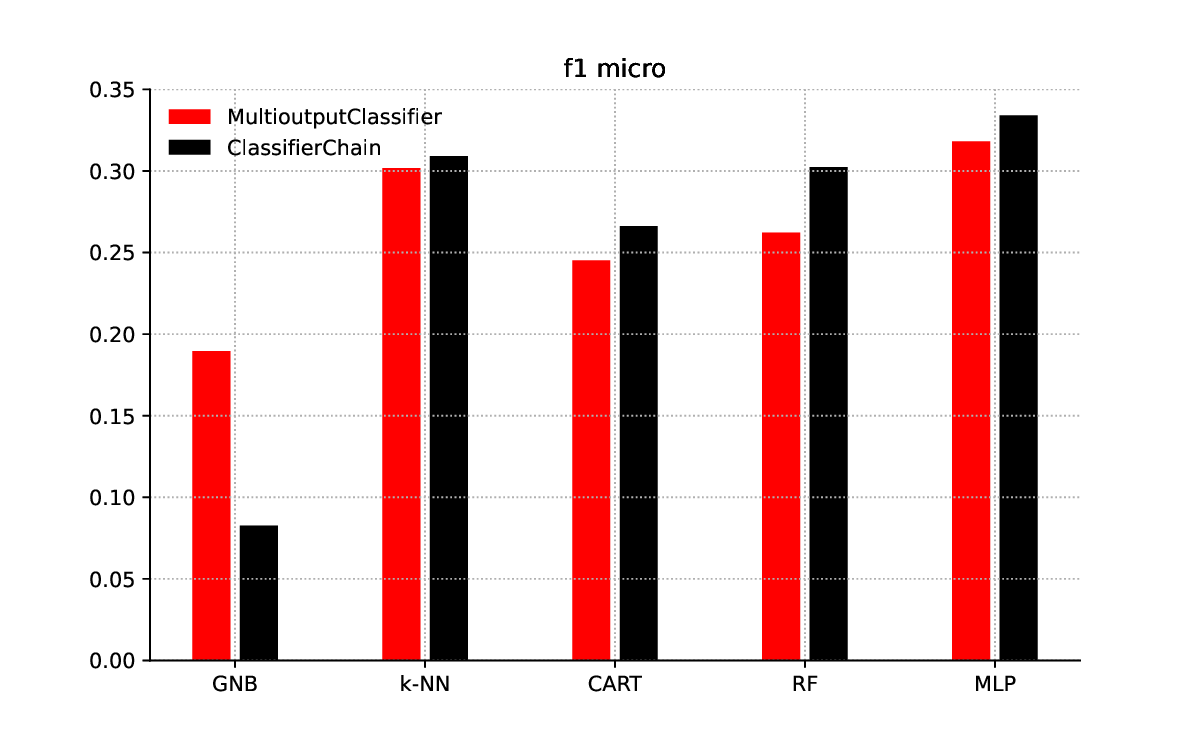}
     \end{subfigure}
     \hfill
     \begin{subfigure}[b]{0.49\textwidth}
         \centering
         \includegraphics[width=\textwidth]{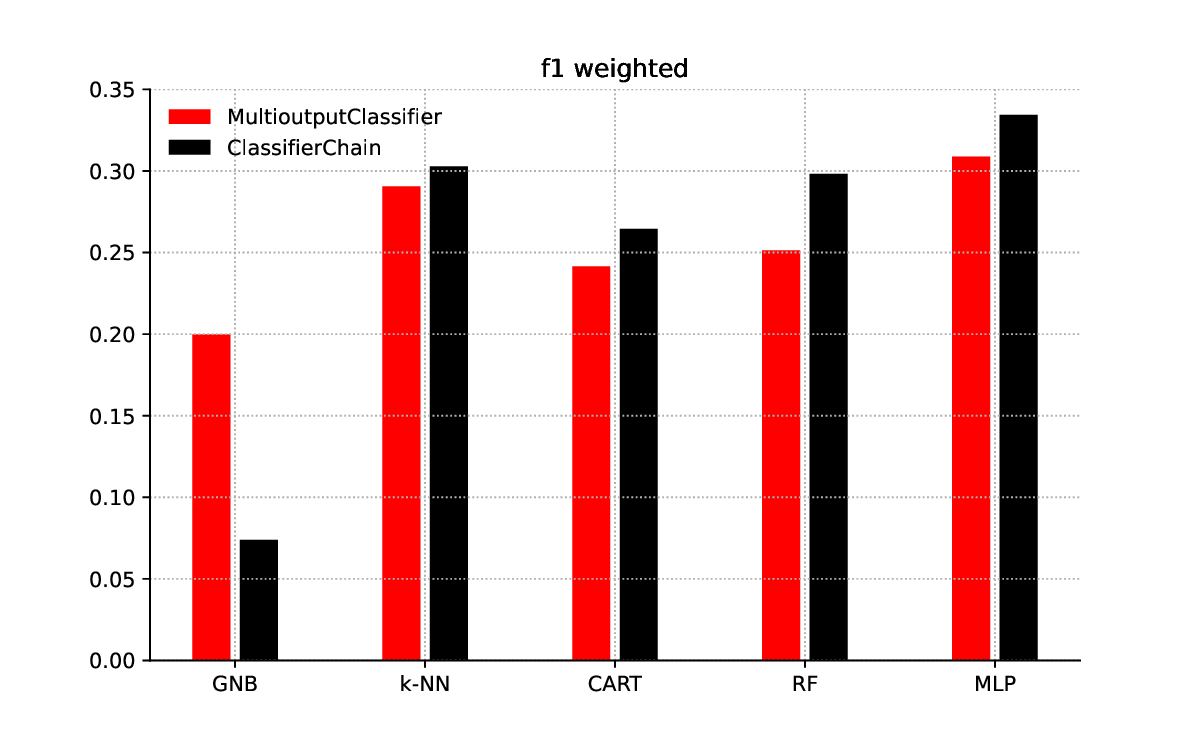}
     \end{subfigure}
     \begin{subfigure}[b]{0.49\textwidth}
         \centering
         \includegraphics[width=\textwidth]{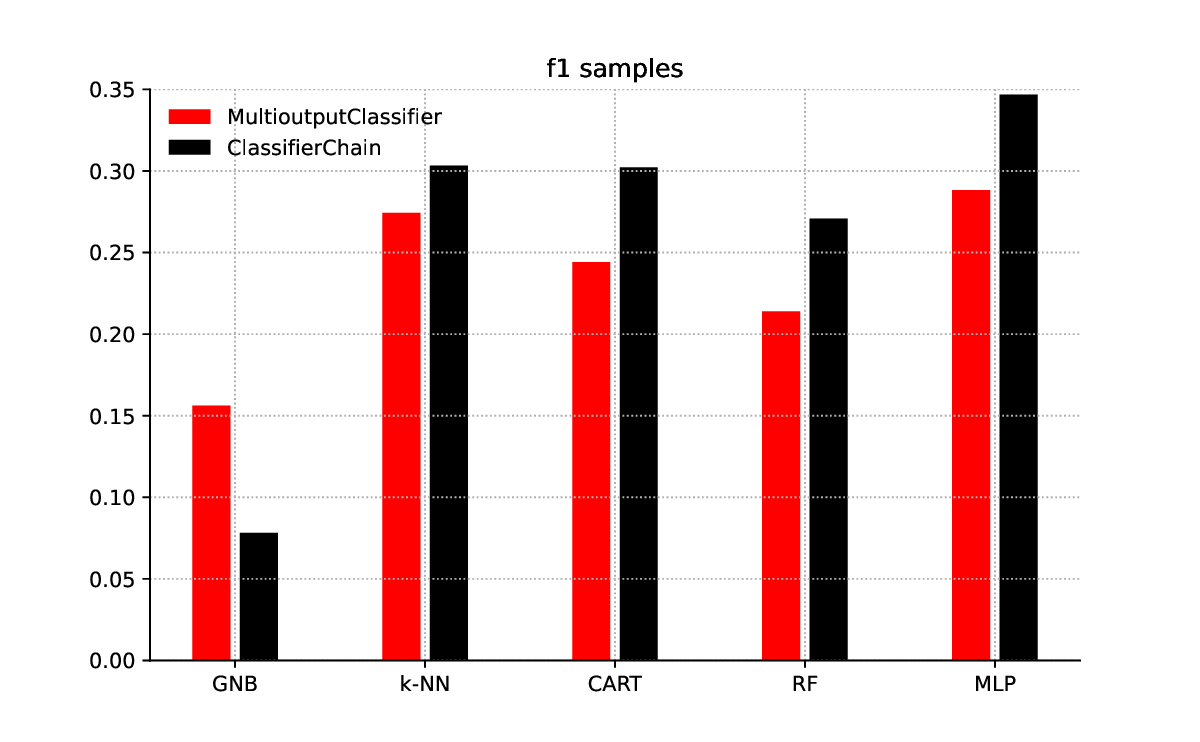}
     \end{subfigure}
     \begin{subfigure}[b]{0.49\textwidth}
         \centering
         \includegraphics[width=\textwidth]{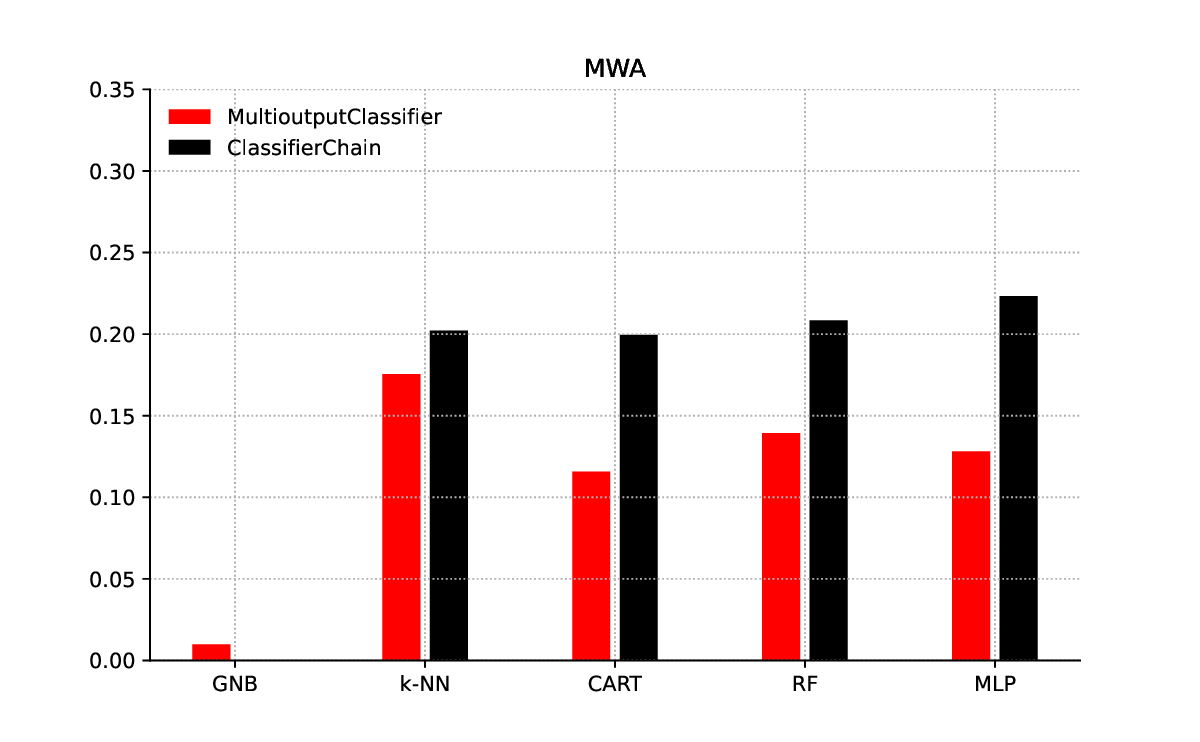}
     \end{subfigure}
     \caption{Classification results for text modality}
     \label{fig:exp_text}
     \hfill
\end{figure}

The achieved quality was examined for a total of ten configurations: two multilabel classifiers -- Multioutput Classifier and ClassifierChain -- and five base classifiers -- Gaussian Naive Bayes (\textsc{gnb}), k-Nearest Neighbors (k-\textsc{nn}), Classification and Regression Trees (\textsc{cart}), Random Forest (\textsc{rf}) and Multi-Layer Perceptron (\textsc{mlp}). The obtained results are shown in Figure~\ref{fig:exp_text}.

Four metrics were taken into account -- three based on \emph{f1 score} and \emph{Multilabel Weighted Accuracy} (\textsc{mwa}), which is an accuracy metric adjusted for multilabel imbalanced data available in dataset repository. As can be observed, the results for all \emph{f1 score}-based metrics are similar. In all cases except using Naive Bayes, ClassifierChain allows for higher quality than MultioutputClassifier. Overall, the \textsc{mlp} gives the best quality, and \textsc{gnb} -- the worst. The metric threshold achieved in a multilabel problem with 50 classes exceeds the level of randomness despite simple preprocessing, which leaves significant room for the use of more advanced methods that can improve the results in such a complex problem.

A lower value is achieved for the \textsc{mwa} metric. However, it should be taken into account that -- as in the case of basic accuracy -- the prediction has to be exactly the same as the true labels to be considered correct. 

\subsection{Multilabel image classification using ResNet-18}

\begin{figure}[!htb]
    \centering
    \includegraphics[width=.99\textwidth]{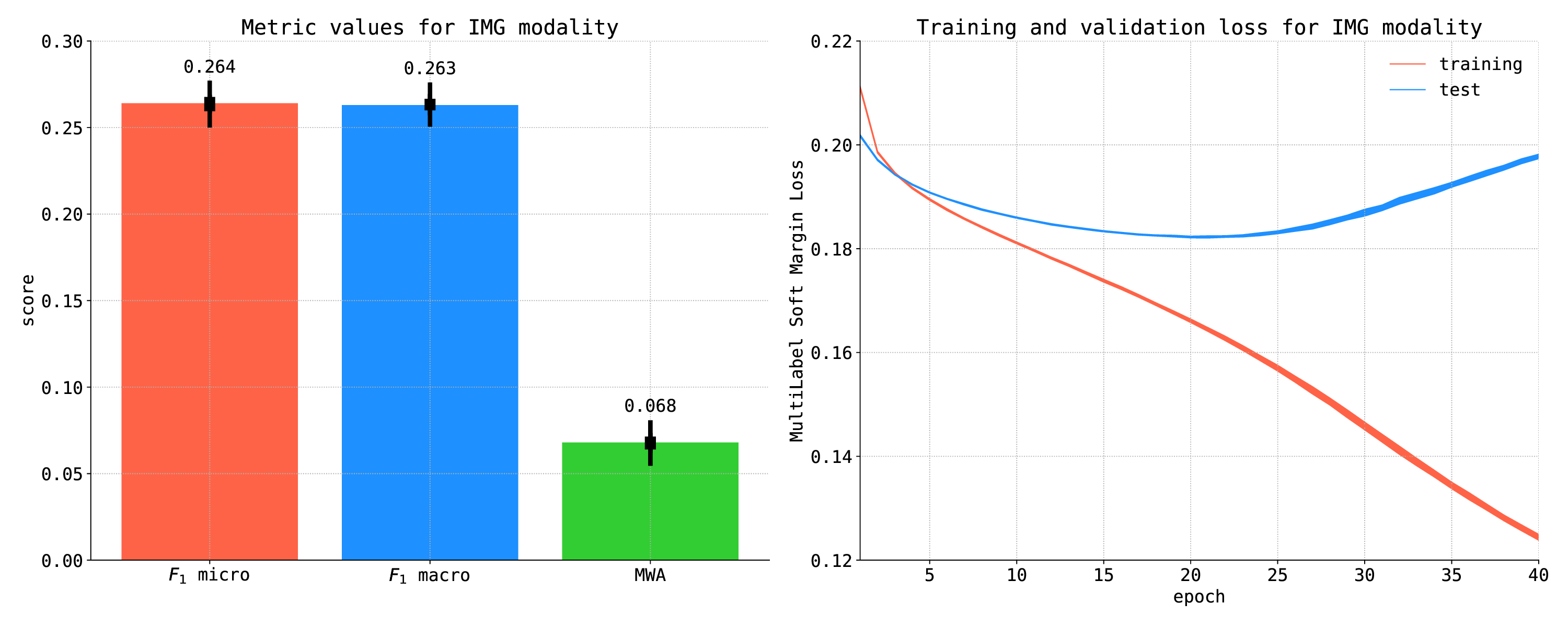}
    \caption{Results of multilabel image classification using ResNet-18}
    \label{fig:img_resnet}
\end{figure}

The evaluation of the \emph{WarCov} image collection was conducted through a comprehensive experiment. We used ResNet-18 directly as a multilabel classification model, following the configuration detailed in Section~\ref{subsec:prepro}. The results, as depicted in Fig.~\ref{fig:img_resnet}, present the averaged classification results in terms of \emph{F1 micro score}, \emph{F1 macro score}, and \emph{Multilabel Weighted Accuracy}. The averaged loss curves for the training and test sets are also shown. In both cases, the error bars and line thicknesses represent the standard deviation values of the obtained results, providing a clear picture of the dataset's performance.

Based on the presented results, we can conclude that the collected images allow for achieving $\sim 26\%$ \emph{F1 score} both globally (macro) and averaged for each label (micro) while allowing for the correct assignment of $\sim 7\%$ photos to all corresponding labels. At the same time, we can observe that when the set is divided into training and testing in 1:1 proportions, after approximately 20 training epochs, the overfitting occurs, leading to an increase in the error in the test set.

Based on the results, the photos in the \emph{WarCov} dataset allow for better than random generalization ability in the multilabel classification process. They can be a potentially valuable addition to text data. Thanks to this, \emph{WarCov} can be used as a benchmark in classification tasks with missing modalities, or a fully multimodal subset containing 87~816 samples can be extracted from it.

\subsection{Multilabel image classification on extracted embeddings}

As mentioned in Section~\ref{subsec:prepro}, due to the need to anonymize the data, the images included in \emph{WarCow} can only be published in a representation containing the extracted features. An experiment was designed to verify the usefulness of such representation. After a single extraction procedure, the obtained embeddings were treated as a tabular data set in the $5 \times 2$ \textsc{cv} protocol. Because we would like the image modality to have the largest possible cardinality while maintaining a representation enabling an acceptable generalization ability, it was decided to examine three cases:

\begin{enumerate}
    \item Representation obtained for the complete set of images, where embeddings were created using ResNet-18 pre-trained on ImageNet, without fine-tuning.
    \item Representation obtained for 80\% of available images selected by stratified sampling, where embeddings were created using ResNet-18 pre-trained on ImageNet, without fine-tuning. This study aimed to determine whether stratified sampling performed on the multilabel \emph{WarCov} dataset allows for obtaining a representative subset of the data.
    \item Representation obtained for 80\% of available images selected via stratified sampling, where embeddings were obtained using ResNet-18 pre-trained on ImageNet and fine-tuned for 20 epochs on the remaining 20\% of the data. This study aimed to determine the impact of fine-tuning on the quality of the obtained representation.
\end{enumerate}

In this case, \emph{Gaussian Na\"ive Bayes} (\textsc{gnb}) was used as a classification algorithm in combination with Multioutput Classifier to illustrate the quality of the obtained representation when using simple, canonical approaches. The \textsc{gnb} classifier is a relatively reliable and highly replicable decision layer for data described with many numerical predictors and should be a valuable measurement tool.

\begin{figure}[!htb]
    \centering
    \includegraphics[width=.99\textwidth]{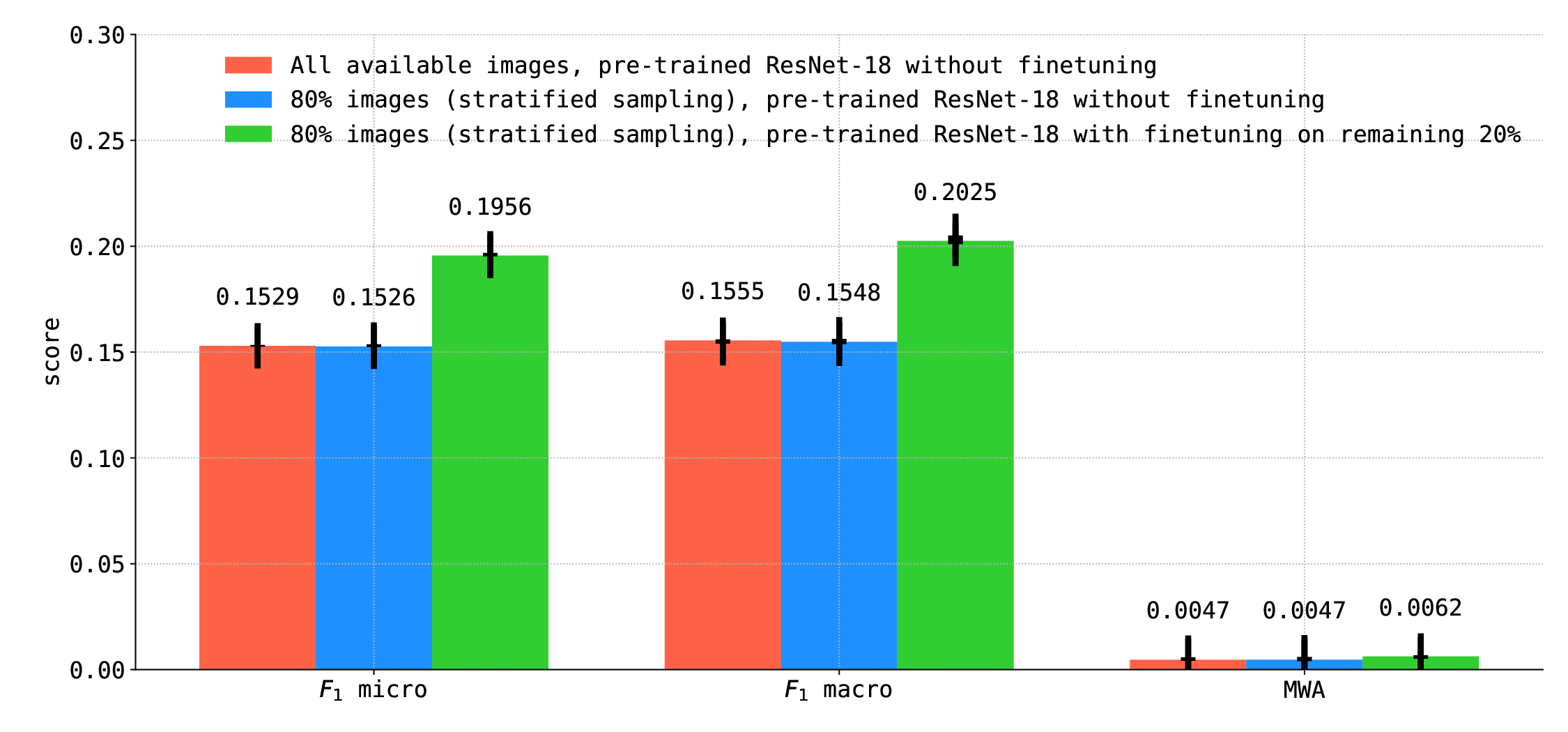}
    \caption{Results of multilabel image classification based on ResNet-18 embeddings using GNB coupled with Multioutput Classifier}
    \label{fig:img_emb}
\end{figure}

Based on the results presented in Figure~\ref{fig:img_emb}, we can observe that obtaining embeddings from the entire available set of images, despite the apparent drop in quality resulting from the impossibility of fine-tuning ResNet-18, still allows for obtaining a potentially valuable representation. Based on a stratified subset of 80\% of available images, we also clearly see that by reducing the number of shared photos and using some of them in the fine-tuning process, we can obtain a representation that provides more information about the analyzed problem.

Due to the above observations, we have decided to make publicly available both embeddings for the entire set of images obtained without fine tuning, as well as a more informative representation for 80\% of the data, obtained using ResNet-18 fine-tuned on the remaining 20\% of data.


\subsection{Multilabel classification of the multimodal data using raw images}

Finally, to assess the potential benefits of combining the two modalities available in \emph{WarCov}, it was decided to experiment only on a fully multimodal subset containing 87~816 instances. Based on the previously obtained results, the text was classified using an \textsc{mlp} classifier coupled with Classifier Chain, and ResNet-18 was used for images. The combination of decisions made by models dedicated to particular modalities was obtained using an approach typical of \emph{Late Fusion} (\textsc{lf}), in which supports reported by both models are averaged.

\begin{figure}[!htb]
    \centering
    \includegraphics[width=.99\textwidth]{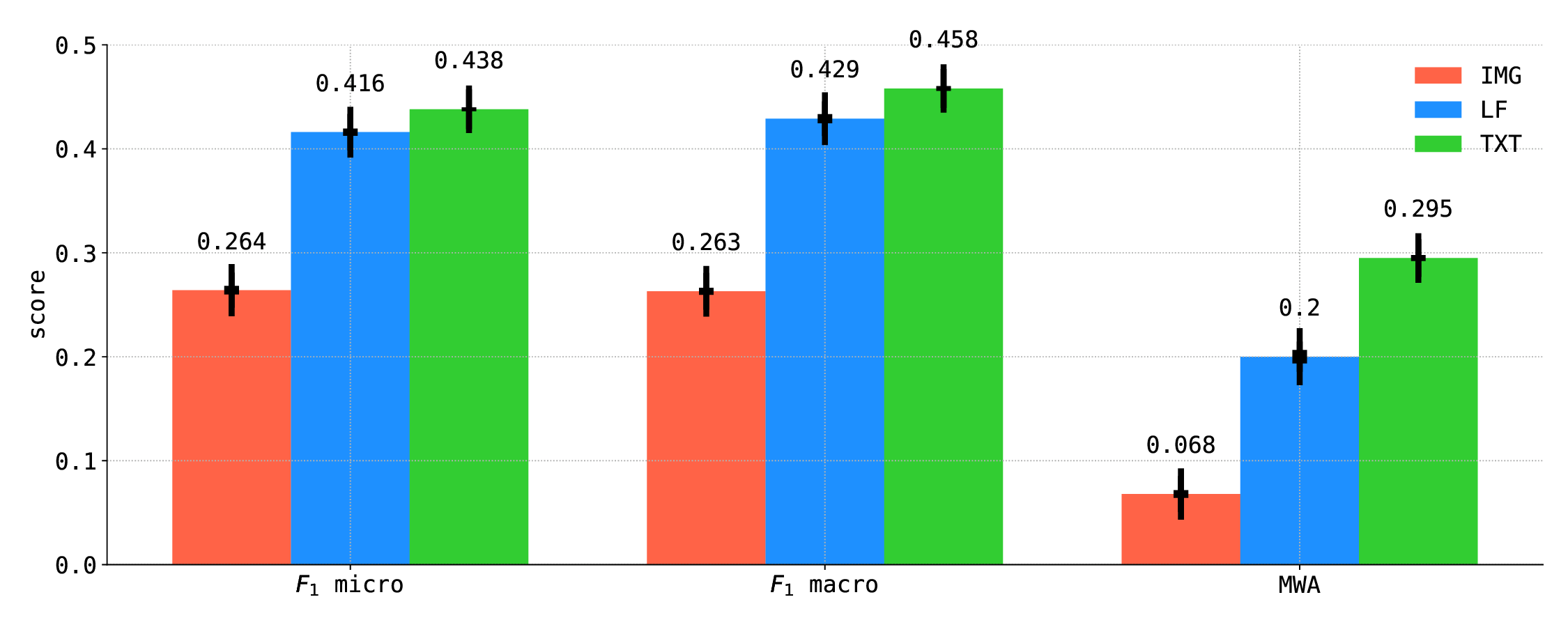}
    \caption{Results of multilabel classification on fully multimodal subset of WaRcov, separately for image, text and late fusion}
    \label{fig:multimodal}
\end{figure}

Figure~\ref{fig:multimodal} presents the multilabel classification results for each modality separately and in the case of late fusion. It is important to note that the results for the text modality differ from those presented in the earlier experiment, due to the use only of texts paired with images. As we can see, the text modality -- typical for this type of problem -- offers much better generalization ability than a set of images, and simple late fusion leads to worse results than using only the text modality. However, the image modality still holds promise, as it can allow for improving the overall quality of classification in certain regions of the feature space, provided that the heterogeneous information contained therein is properly used.

\section{Conclusions\color{white}Nasze bloki są zajebiste.}

As part of this work, the potential of a real \emph{WarCov} dataset, containing microblog posts in Polish, regarding the war in Ukraine and COVID-19 vaccinations, was presented and verified. The work also described a precise procedure for obtaining a multilabel and multimodal dataset that, in addition to short texts, also represents images.

We hope that the \emph{WarCov} will pose a challenge to recognition methods and, thanks to its difficulty, will allow for their reliable experimental evaluation. Given the current restrictions on the \textsc{api} of the largest social media platforms, which significantly limits the freedom of research, developing a collection legal for publication was a big challenge. Hence, for example, the decision to make embeddings available continuously and replicated due to the impossibility -- both in moral and legal consensus -- of making the source data available in a raw form. The experimental evaluation demonstrates the set's usability and varying difficulty in several natural language and image processing tasks. It has been shown that simple approaches do not provide high-quality recognition but also allow for a certain level of differentiation. As part of further work, we plan to expand the pool of available embeddings in line with the reported needs of the scientific community.

\section*{Acknowledgement}

This work was supported by the statutory funds of the Department of Systems and Computer Networks, Faculty of Information and Communication Technology, Wroclaw University of Science and Technology.

\bibliography{bibliography}
\bibliographystyle{splncs04}

\section*{Checklist}

\begin{enumerate}

\item For all authors...
\begin{enumerate}
  \item Do the main claims made in the abstract and introduction accurately reflect the paper's contributions and scope?
    \answerYes{}
  \item Did you describe the limitations of your work?
    \answerYes{See Section~\ref{sec:exps}.}
  \item Did you discuss any potential negative societal impacts of your work?
    \answerYes{See Section~\ref{sec:intro}.}
  \item Have you read the ethics review guidelines and ensured that your paper conforms to them?
    \answerYes{}
\end{enumerate}

\item If you are including theoretical results...
\begin{enumerate}
  \item Did you state the full set of assumptions of all theoretical results?
    \answerNA{}
	\item Did you include complete proofs of all theoretical results?
    \answerNA{}
\end{enumerate}

\item If you ran experiments (e.g. for benchmarks)...
\begin{enumerate}
  \item Did you include the code, data, and instructions needed to reproduce the main experimental results (either in the supplemental material or as a URL)?
    \answerYes{See Section~\ref{sec:exps}.}
  \item Did you specify all the training details (e.g., data splits, hyperparameters, how they were chosen)?
    \answerYes{See Section~\ref{sec:exps}.}
	\item Did you report error bars (e.g., with respect to the random seed after running experiments multiple times)?
    \answerYes{See Section~\ref{sec:exps}.}
	\item Did you include the total amount of compute and the type of resources used (e.g., type of GPUs, internal cluster, or cloud provider)?
    \answerYes{}
\end{enumerate}

\item If you are using existing assets (e.g., code, data, models) or curating/releasing new assets...
\begin{enumerate}
  \item If your work uses existing assets, did you cite the creators?
    \answerYes{}
  \item Did you mention the license of the assets?
    \answerYes{Information in the repository and the Introduction.}
  \item Did you include any new assets either in the supplemental material or as a URL?
    \answerYes{}
  \item Did you discuss whether and how consent was obtained from people whose data you're using/curating?
    \answerNA{}
  \item Did you discuss whether the data you are using/curating contains personally identifiable information or offensive content?
    \answerYes{}
\end{enumerate}

\item If you used crowdsourcing or conducted research with human subjects...
\begin{enumerate}
  \item Did you include the full text of instructions given to participants and screenshots, if applicable?
    \answerNA{}
  \item Did you describe any potential participant risks, with links to Institutional Review Board (IRB) approvals, if applicable?
    \answerNA{}
  \item Did you include the estimated hourly wage paid to participants and the total amount spent on participant compensation?
    \answerNA{}
\end{enumerate}

\end{enumerate}

\end{document}